\documentclass{article}

\usepackage[nonatbib, final]{nips_2016}

\usepackage[utf8]{inputenc} 
\usepackage[T1]{fontenc}    
\usepackage[pdfborder={0 0 0}]{hyperref}       
\usepackage{url}            
\usepackage{booktabs}       
\usepackage{amsfonts}       
\usepackage{nicefrac}       
\usepackage{microtype}      
\usepackage{amsmath}		
\usepackage{graphicx}
\usepackage{caption}
\usepackage{subcaption}
\usepackage{authblk}

\title{Voxelwise nonlinear regression toolbox for neuroimage analysis: Application to aging and neurodegenerative disease modeling}

\author[1]{Santi Puch}
\author[1]{Asier Aduriz}
\author[1]{Adrià Casamitjana}
\author[1]{Veronica Vilaplana}
\author[2]{Paula Petrone}
\author[2]{Grégory Operto}
\author[2]{Raffaele Cacciaglia}
\author[2]{Stavros Skouras}
\author[2]{Carles Falcon}
\author[2]{José Luis Molinuevo}
\author[2]{Juan Domingo Gispert} 

\affil[1]{Universitat Politècnica de Catalunya, Barcelona, Spain \\
\texttt{veronica.vilaplana@upc.edu}}
\affil[2]{Barcelona$\beta$eta Brain Research Center, Barcelona, Spain}

\begin{document}

\maketitle

\begin{abstract}
This paper describes a new neuroimaging analysis toolbox that allows for the modeling of nonlinear effects at the voxel level, overcoming limitations of methods based on linear models like the GLM. We illustrate its features using a relevant example in which distinct nonlinear trajectories of Alzheimer's disease related brain atrophy patterns were found across the full biological spectrum of the disease. 

The open-source toolbox is available in GitHub: \url{https://github.com/imatge-upc/VNeAT}.
\end{abstract}

\section{Introduction}

Nowadays there is a vast armory of neuroimaging analysis tools available for the neuroscientific community whose ultimate goal is to conduct spatially extended statistical tests to identify regionally significant effects in the images without any a priori hypothesis on the location or extent of these effects. Some of these tools perform these tests at the voxel level\footnote{\url{http://www.fil.ion.ucl.ac.uk/spm/}}, whereas some others provide with specific environments for analyzing the images, based on 3D meshes\footnote{\url{http://freesurfer.net/}} or boundaries\footnote{\url{http://idealab.ucdavis.edu/software/bbsi.php}}. 

Irrespective of these differences, the vast majority of neuroimaging tools are based on different implementations of the General Linear Model (GLM). Even though the GLM is flexible enough for conducting most of the typical statistical analysis, its flexibility for modeling nonlinear effects is rather limited. To this regard, it is important to note that some relevant confounders in most analysis, such as the impact of age on regional brain volumes, have been found to be better described by nonlinear processes \cite{nonlinear_subcortical,nonlinear_cortical}.

In this event, the most widely used alternative is to refrain from performing a voxel-wise analysis and quantify brain volumes based on regions of interest (ROIs) and then perform the nonlinear fitting of these trajectories by external software. This alternative has the obvious disadvantage that made voxel-wise analysis so popular on the first place: the need to identify a set of ROIs a priori. 

In this work, we describe a new analysis toolbox that allows for the modeling of nonlinear effects at the voxel level that overcomes these limitations. In the following sections we briefly describe the main functionalities of the toolbox and illustrate its features using a relevant example in which distinct nonlinear trajectories are found as a function of a cerebrospinal fluid (CSF) related biomarker for participants in all stages of Alzheimer's disease (AD).

\section{The toolbox}
\label{toolbox}

The toolbox comprises an independent fitting library, made up of different \textit{model fitting} and \textit{fit evaluation} methods, a processing module that interacts with the fitting library providing the formatted data obtained from the file system, several \textit{visualization} tools and a command line interface that allows the interaction between the user and the processing module, supported by a configuration file. 

\subsection{Model fitting techniques}

Model fitting consists in finding a parametric or a nonparametric function of some explanatory variables (predictors) and possibly some confound variables (correctors) that best fits the observations of the target variable in terms of a given quality metric or, conversely, that minimizes the loss between the prediction of the model and the actual observations. The models included in this toolbox are:

\textbf{General Linear Model (GLM)} 

The General Linear model is a generalization of multiple linear regression to the case of more than one dependent variable. Nonlinear relationships can be modeled within the GLM framework using a polynomial basis expansion, mapping the input space into a feature space that includes the polynomial terms of the variables.

\textbf{Generalized Additive Model (GAM)} 

A Generalized Additive Model \cite{gam_Hastie_1990} is a Generalized Linear Model in which the observations of the target variable depend linearly on unknown smooth functions of some predictor variables: $ f(X) = \alpha + \sum_{i=1}^{k} f_i(X_i)$. Here $f_1, f_2, ..., f_k$ are nonparametric smooth functions that are simultaneously estimated using scatterplot smoothers by means of the backfitting algorithm \cite{breiman_backfit}. Several fitting methods can be accommodated in this framework by using different smoother operators, such as cubic splines, polynomial or Gaussian smoothers. 

\textbf{Support Vector Regression (SVR)} 

In Support Vector Regression the goal is to find a function that has at most $\epsilon$ deviation from the observations and, at the same time, is as flat as possible. However, the $\epsilon$ deviation constraint is not feasible sometimes, and a hyperparameter $C$ that controls the degree up to which deviations larger than $\epsilon$ are tolerated is introduced. 
In the context of SVR nonlinearities are introduced with the "kernel trick", that is, a kernel function $ k(x_i, x_j) = \langle \Phi(x_i), \Phi(x_j) \rangle $ that implicitly maps the inputs from their original space into a high-dimensional space. The kernel function used in this toolbox is the Radial Basis Function (RBF) kernel, which is defined as $ k(x_i, x_j) = exp(-\gamma \|x_i - x_j\|^2)$.

SVR methods rely on several hyperparameters, namely $\epsilon$ and $C$ in general and also $\gamma$ when using a RBF kernel function. To address the search of these hyperparameters an automatic method based on grid search is included in this toolbox, which comprises the following steps: 1) sample the hyperparameters space in a grid using one of the several sampling methods provided in the toolbox; 2) fit a subset of the data with the combination of hyperparameters of each sample in the grid; 3) select the combination that minimizes the error function of choice. 
The lack of validation data and the nature of morphometric data, which is vastly dominated by voxels with 0-valued observations, limits the ability to find a subset of the data that is valid to find the optimal hyperparameters without incurring in overfitting or underfitting. For that reason the following approach is taken: a subset of $m$ voxels that contain observations with a minimum variance of $Var_{min}$ is selected in each of the $N$ iterations of the algorithm, and the error computed for each voxel is weighted by the inverse of the variance of its observations. 

\subsection{Fit evaluation methods}

The goodness of fit of a model can be evaluated using different metrics in order to create 3D statistical maps, as shown in \autoref{fig:ftest_svrpol}.
		
\textbf{MSE}, \textbf{Coefficient of determination} ($\mathbf{R^2}$)

These two metrics evaluate the predictive power of a model without penalizing its complexity.

\textbf{Akaike Information Criterion (AIC)}

The AIC is a criterion based on information theory widely used for model comparison and selection. Unlike the previous metrics, it penalizes the complexity of the model by requiring its number of parameters.

\textbf{F-test} 

The F-test evaluates whether the variance of the full model (correctors and predictors) is significantly lower — from a statistical point of view — than the variance of the restricted model (only correctors), that is, the inclusion of the predictors contributes to the explanation of the observations. 
This statistical test requires the degrees of freedom of both models, which are trivial to compute in GLM, but are not in GAM or SVR. The equivalent degrees of freedom for Support Vector Regression are introduced in the toolbox as defined in \cite{equivalent_df_SVR}.

\textbf{Penalized Residual Sum of Squares (PRSS), Variance-Normalized PRSS} 

Penalized Residuals Sum of Squares is introduced in the toolbox in order to provide a fit evaluation metric that penalizes the complexity of the predicted curve without requiring the degrees of freedom. However, PRSS is not suitable enough in the context of morphometric analysis, as it always provides better scores for target variables with low-variance and flat trends than for target variables with high-variance and non-flat trends, and that poses a problem as most of the voxels in the brain are 0-valued.
For that reason a variance normalized version of the PRSS that weights the score with the inverse of the variance of the predicted curve, the Variance Normalized Penalized Residual Sum of Squares, has also been implemented in the toolbox.

\subsection{Model comparison and Interactive visualization tool}

The toolbox provides various methods to compare statistical maps generated using different fitting models. These methods can be used, for instance, to select the best fitting model for each voxel, to visualize the relative contribution of different models or to validate the similarity between two statistical maps.

Moreover, an interactive visualization tool is included to give additional insight on the results: it allows to load a 3D statistical map — generated with the fit evaluation method of choice or with a comparison method — and one or several fitted models, and then plot the predicted curves of all the models for the voxel selected with the cursor, hence easing the task of inspecting the curves in the significant regions. An example of the aforementioned tool is found in \autoref{fig:gsvr_best}. Here, a best-fit map is shown, where voxels are labeled according to the best fit score of three models under comparison (polynomial GLM, polynomial SVR and Gaussian SVR). The bottom-right plot shows the trajectories obtained for the three methods for the selected voxel and the corrected observations.

\section{Implementation details}

The whole toolbox has been implemented in Python 2.7. NumPy and SciPy libraries have been used for the numerical and scientific computing, NiBabel to handle the morphometric data in NIfTI format, scikit-learn (\cite{scikit-learn}) for the machine learning algorithms and matplotlib and seaborn for plotting and the visualization features.

\section{A case study: atrophy patterns across the Alzheimer's disease continuum}

To illustrate the toolbox functionalities, we analyze the nonlinear volumetric changes in gray matter across the AD's spectrum, following the approach proposed in \cite{Nonlinear_Gispert_2015}. Here, the subject's level of pathology and position along the AD continuum is represented by a CSF-related biomarker, the AD-CSF index introduced in \cite{ADCSF_Molinuevo_2013}. 

The dataset comprises 129 participants (62 controls, 18 preclinical AD, 28 mild cognitive impairment (MCI) due to AD and 21 with diagnosed AD) who underwent MRI scanning and CSF analysis. AD-CSF index was computed for all subjects applying the formula in \cite{ADCSF_Molinuevo_2013}, and the T1-weighted brain scans of all participants were pre-processed using Voxel Based Morphometry (VBM) as implemented in SPM8\footnote{\url{http://www.fil.ion.ucl.ac.uk/spm}}. Further details of the 
dataset and the image processing pipeline can be found in \cite{Nonlinear_Gispert_2015}.

Images of all the subjects were pooled together regardless of their diagnostic classification. We first applied the same model proposed in \cite{Nonlinear_Gispert_2015}, a polynomial GLM, entering age, sex, and the AD-CSF index as regressors, modeling age as a second order polynomial to correct for the effects of aging on gray matter content and the AD-CSF index as a third order polynomial. F-tests were used to analyze the significant effects regarding the three AD-CSF index terms. Results were coherent with the ones found in \cite{Nonlinear_Gispert_2015}.

\begin{figure}[!htp]
	\centering
	\begin{minipage}{.6\textwidth}
		\captionsetup{justification=centering,margin=0.5cm}
		\centering
		\includegraphics[width=.95\linewidth]{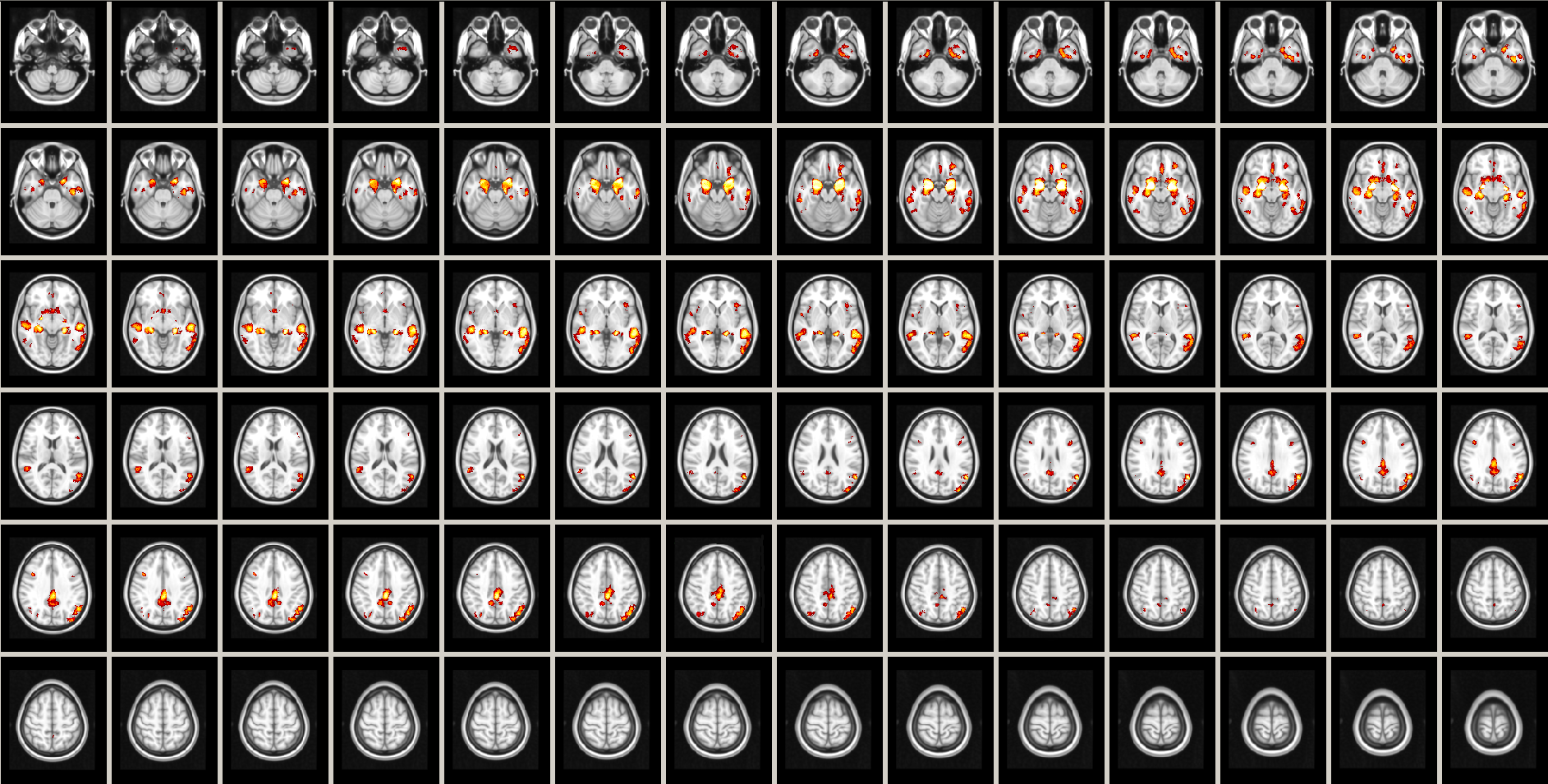}
		\captionof{figure}{F-test statistical map for the polynomial SVR, with significance level ($\alpha$) filtering at 0.001, minimum cluster size of 100 voxels and transformed into Z-scores for improved
		visualization.}
		\label{fig:ftest_svrpol}
	\end{minipage}
	\begin{minipage}{.35\textwidth}
		\centering
		\captionsetup{justification=centering,margin=0.1cm}
		\includegraphics[width=.95\linewidth]{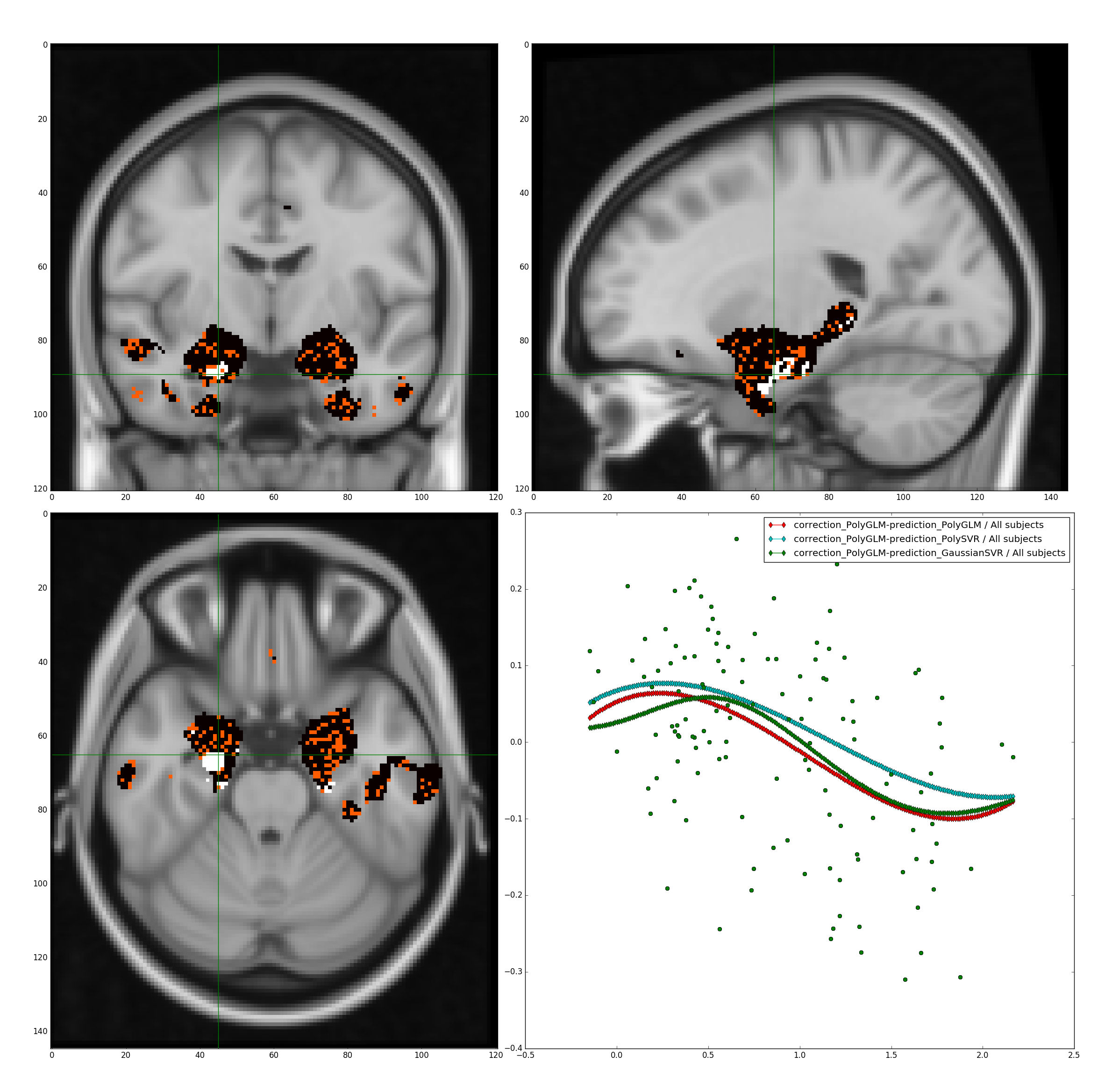}
		\captionof{figure}{Comparison of polynomial GLM (black) vs polynomial SVR (orange) vs Gaussian SVR (white) with the corresponding curves for a voxel in which the best fitting score belongs to the Gaussian SVR model.}
		\label{fig:gsvr_best}
	\end{minipage}%
\end{figure}

In a second experiment we fitted a polynomial and a Gaussian SVR and computed their F-test statistical maps to compare their behavior with the polynomial GLM. As an example, the map corresponding to the polynomial SVR is shown in \autoref{fig:ftest_svrpol}.

To compare the models, a best-fit map was obtained, assigning different numerical labels to voxels (label $i$ is assigned if the best fit is for the $i$th model).  Only voxels at a significance level of  $p<0.001$ located within clusters of more than 100 voxels are shown. An example of a voxel where the best score corresponds to the Gaussian SVR is presented in \autoref{fig:gsvr_best}, using the curve visualization tool.

\section{Conclusions}

As compared to ROI-based statistical analysis, the voxelwise approach has the key advantage to allow for spatially unbiased analysis of brain images. This strategy has become predominant in the last decades and typically relies on particular implementations of the GLM. However, this approach presents limitations when it comes to the modeling of nonlinear effects, hence current neuroimaging analysis tools are sub-optimal for the identification of such nonlinear patterns. As a consequence, the capacity of neuroscientists to detect spatially distributed critical points associated to brain maturation or pathology is as well limited. In this context we understand that nonlinear modeling tools like the one that we are describing in this paper could be interesting and useful for the neuroimaging community. 

As future work, we plan to extend the toolbox to deal with longitudinal data. The proliferation of neuroimaging repositories with longitudinal data provides with a unique opportunity for imaging researchers to analyze reference datasets with different tools, thus enabling the comparison and validation of analytical tools while gaining more insight on the data under analysis. Therefore, in the analysis of such longitudinal datasets, the availability of nonlinear modeling tools is likewise critical to fully understand the interrelationships between the trajectories of different biomarkers that may be crucial for understanding downstream pathological effects.

The open-source toolbox is available in GitHub: \url{https://github.com/imatge-upc/VNeAT}.



\section*{Acknowledgements}
This work has been developed in the framework of the project BigGraph TEC2013-43935-R, funded by the Spanish Ministerio de Economia y Competitividad and the European Regional Development Fund (ERDF). Adrià Casamitjana is supported by the
Spanish Ministerio de Educación, Cultura y Deporte FPU Research Fellowship.

\small

\nocite{statistics_Hastie_2009}
\nocite{tutorial_SVR}

\bibliographystyle{unsrt}
\bibliography{bibliography}

\begin{thebibliography}{10}

\bibitem{nonlinear_subcortical}
A.~M. Fjell, L.~T. Westlye, H.~Grydeland, I.~Amlien, T.~Espeseth, I.~Reinvang,
  N.~Raz, D.~Holland, A.~M. Dale, and K.~B. Walhovd.
\newblock {{C}ritical ages in the life course of the adult brain: nonlinear
  subcortical aging}.
\newblock {\em Neurobiol. Aging}, 34(10):2239--2247, Oct 2013.

\bibitem{nonlinear_cortical}
A.~M. Fjell, L.~T. Westlye, H.~Grydeland, I.~Amlien, T.~Espeseth, I.~Reinvang,
  N.~Raz, A.~M. Dale, and K.~B. Walhovd.
\newblock {{A}ccelerating cortical thinning: unique to dementia or universal in
  aging?}
\newblock {\em Cereb. Cortex}, 24(4):919--934, Apr 2014.

\bibitem{gam_Hastie_1990}
Trevor~J Hastie and Robert~J Tibshirani.
\newblock {\em Generalized Additive Models}, volume~43 of {\em Monographs on
  Statistics and Applied Probability}.
\newblock Chapman \& Hall/CRC, June 1990.

\bibitem{breiman_backfit}
L.~Breiman and J.H. Friedman.
\newblock {{E}stimating optimal transformations for multiple regression and
  correlations (with discussion)}.
\newblock {\em Journal of the American Statistical Association},
  80(391):580--619, 1985.

\bibitem{equivalent_df_SVR}
Francesco Dinuzzo, Marta Neve, Giuseppe~De Nicolao, and Ugo~Pietro Gianazza.
\newblock On the representer theorem and equivalent degrees of freedom of svr.
\newblock {\em Journal of Machine Learning Research}, 8:2467--2495, December
  2007.

\bibitem{scikit-learn}
F.~Pedregosa, G.~Varoquaux, A.~Gramfort, V.~Michel, B.~Thirion, O.~Grisel,
  M.~Blondel, P.~Prettenhofer, R.~Weiss, V.~Dubourg, J.~Vanderplas, A.~Passos,
  D.~Cournapeau, M.~Brucher, M.~Perrot, and E.~Duchesnay.
\newblock Scikit-learn: Machine learning in {P}ython.
\newblock {\em Journal of Machine Learning Research}, 12:2825--2830, 2011.

\bibitem{Nonlinear_Gispert_2015}
Juan~Domingo Gispert, Lorena Rami, G~S{\'a}nchez-Benavides, C~Falcon, Alan
  Tucholka, S~Rojas, and Jose~L Molinuevo.
\newblock Nonlinear cerebral atrophy patterns across the alzheimer's disease
  continuum: impact of apoe4 genotype.
\newblock {\em Neurobiology of Aging}, 36(10):2687--2701, October 2015.

\bibitem{ADCSF_Molinuevo_2013}
Jose~L Molinuevo, Juan~Domingo Gispert, Bruno Dubois, Michael Heneka, Alberto
  Lleo, Sebastiaan Engelborghs, Jes{\'u}s Pujol, Leonardo~Cruz de~Souza, Daniel
  Alcolea, Frank Jessen, Marie Sarazin, Foudil Lamari, Mircea Balasa, Anna
  Antonell, and Lorena Rami.
\newblock The ad-csf-index discriminates alzheimer’s disease patients from
  healthy controls: a validation study.
\newblock {\em Journal of Alzheimer's Disease}, 36(1):67--77, June 2013.

\bibitem{statistics_Hastie_2009}
Trevor Hastie, Robert Tibshirani, and Jerome Friedman.
\newblock {\em The Elements of Statistical Learning. Data Mining, Inference,
  and Prediction}.
\newblock Springer, second edition edition, February 2009.

\bibitem{tutorial_SVR}
Alex~J. Smola and Bernhard Sch\"{o}lkopf.
\newblock A tutorial on support vector regression.
\newblock {\em Statistics and Computing}, 14(3):199--222, August 2004.

\end{thebibliography}

\end{document}